\documentclass[letterpaper, 10 pt, conference]{ieeeconf}  

\IEEEoverridecommandlockouts                              

\usepackage{cite}
\usepackage{amsmath,amssymb,amsfonts}
\usepackage{multirow}
\usepackage{float}
\usepackage{subfigure}
\usepackage{comment}
\usepackage{algorithm}
\usepackage{algpseudocode}
\usepackage{booktabs} 
\usepackage{array}    
\usepackage{siunitx}  
\usepackage{graphicx} 
\usepackage[table]{xcolor} 
\usepackage{hyperref}

\hypersetup{
    colorlinks=true,    
    linkcolor=blue,     
    filecolor=magenta,  
    urlcolor=cyan,      
    citecolor=green,    
}

\title{\LARGE \bf
MHRC: Closed-loop Decentralized \underline{M}ulti-\underline{H}eterogeneous \underline{R}obot \underline{C}ollaboration with Large Language Models
}

\author{Wenhao Yu$^{1}$, Jie Peng$^{2}$, Yueliang Ying$^{3}$, Sai Li$^{4}$, Jianmin Ji$^{4, *}$ and Yanyong Zhang$^{2}$
\thanks{$^{1}$ Institute of Advanced Technology, University of Science and Technology of China (USTC), Hefei 230026, China
        {\tt\small wenhaoyu@mail.ustc.edu.cn}}%
\thanks{$^{2}$ School of Artificial Intelligence and Data Science, USTC, 230026, China}%
\thanks{$^{3}$ The University of North Carolina at Chapel Hill, USA}
\thanks{$^{4}$ School of Computer Science and Technology, USTC, 230026, China}%
\thanks{${*}$ Corresponding author. {\tt\small jianmin@ustc.edu.cn}}
}

\begin{document}

\maketitle
\thispagestyle{empty}
\pagestyle{empty}

\begin{abstract}

The integration of large language models (LLMs) with robotics has significantly advanced robots' abilities in perception, cognition, and task planning. The use of natural language interfaces offers a unified approach for expressing the capability differences of heterogeneous robots, facilitating communication between them, and enabling seamless task allocation and collaboration. Currently, the utilization of LLMs to achieve decentralized multi-heterogeneous robot collaborative tasks remains an under-explored area of research. In this paper, we introduce a novel framework that utilizes LLMs to achieve decentralized collaboration among multiple heterogeneous robots. Our framework supports three robot categories—mobile robots, manipulation robots, and mobile manipulation robots—working together to complete tasks such as exploration, transportation, and organization.
We developed a rich set of textual feedback mechanisms and chain-of-thought (CoT) prompts to enhance task planning efficiency and overall system performance. The mobile manipulation robot can adjust its base position flexibly, ensuring optimal conditions for grasping tasks. The manipulation robot can comprehend task requirements, seek assistance when necessary, and handle objects appropriately. Meanwhile, the mobile robot can explore the environment extensively, map object locations, and communicate this information to the mobile manipulation robot, thus improving task execution efficiency.
We evaluated the framework using PyBullet, creating scenarios with three different room layouts and three distinct operational tasks. We tested various LLM models and conducted ablation studies to assess the contributions of different modules. The experimental results confirm the effectiveness and necessity of our proposed framework.

\end{abstract}

\section{INTRODUCTION}

The complexity of real-world tasks poses significant challenges in designing a single robot capable of handling all aspects of task execution. This has driven a growing body of research focused on multi-heterogeneous robot collaboration~\cite{rizk2019cooperative}.
These systems often consist of robots with varying capabilities, such as manipulation and navigation, working collaboratively to achieve common objectives~\cite{shorinwa2024distributed}. In this task context, traditional approaches often rely on explicit programming and centralized control, which tend to be inflexible and inefficient in managing the complexities of real-world scenarios~\cite{kolling2006multirobot}.
 
Recent developments in LLMs offer a promising avenue for enhancing robot collaboration through natural language understanding and generation. Natural language provides a unified interface for robot collaboration, enabling robots to better understand each other's actions and communicate more conveniently.
Recent advancements in LLMs have demonstrated remarkable capabilities in natural language understanding, dialogue generation, extensive world knowledge, and complex reasoning. Leveraging these strengths offers new opportunities to address challenges in decentralized multi-heterogeneous robot collaboration~\cite{liu2024leveraging, mandi2024roco}.

\begin{figure}[t]
    \centering
    \includegraphics[width=\columnwidth]{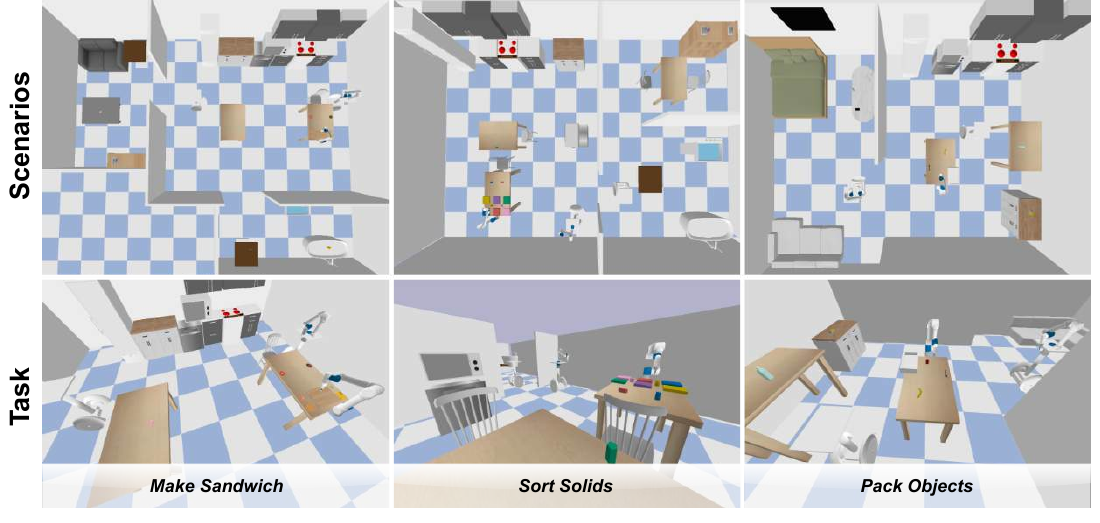}
    \caption{This figure depicts the experimental scenarios and tasks involved in our research. The furniture types and layouts differ across various settings, with distinct room configurations, such as kitchens, bathrooms, and bedrooms. The task design is inspired by RoCo~\cite{mandi2024roco}, encompassing activities such as sandwich making, sorting solid objects, and packing items.}
    \label{fig:first}
\end{figure}

Therefore, we introduce \texttt{MHRC}, a novel framework that harnesses the power of LLMs to enable decentralized multi-heterogeneous robot decision-making and planning. We model the collaborative task as a decentralized partially observable Markov decision process (DEC-POMDP)~\cite{amato2013decentralized,bernstein2013cdcm}, which allows each robot to make decisions based on its local observations and communications with other robots. 
We meticulously designed prompt templates for the tasks, leveraging in-context learning to tap into the extensive world knowledge of LLMs. Furthermore, we employed the CoT~\cite{wei2023cot} framework to enhance the LLMs' capabilities in processing long sequences and performing complex reasoning.

Our proposed framework consists of three key modules:
\textbf{Observation Module}: This module gathers all necessary state information for task and motion planning. It includes a scene graph representing structured information about the environment, messages exchanged between robots, and individual robot status information. The scene graph is dynamically updated to reflect changes in the environment, such as the positions of objects and the states of manipulable furniture. The scene graph updates are based on the robot's local observations and the received messages. Communication between robots is handled through structured natural language prompts, allowing them to share information and request assistance effectively. 
\textbf{Memory Module}: To handle long-horizon tasks and maintain context across multiple interactions, the memory module records feedback history, received message history, and action history. By marking recent feedback and messages with special tags, the module ensures that the most relevant information influences the robots' decision-making processes. 
\textbf{Planning Module}: Utilizing LLMs, each robot independently generates actions by selecting from a predefined list, guided by the CoT framework. After executing an action, the robot receives feedback and updates its memory, allowing it to adjust its plan accordingly. This iterative process enables robots to adapt to new information and collaborate effectively to complete complex tasks.

We validate our framework through experiments conducted in simulated environments created using PyBullet~\cite{greff2022kubric}. The environments feature various room layouts, furniture arrangements, and object distributions to simulate realistic scenarios. We design three distinct tasks—Pack Objects, Sort Solids, and Make Sandwich—to assess the robots' capabilities in basic manipulation, color matching, and sequential stacking, respectively. Each task requires collaboration among three types of robots: a mobile robot responsible for exploration, a mobile manipulation robot that opens furniture and transports objects, and a manipulation robot that performs tabletop operations.

Our evaluation metrics include success rate, partial success rate, average temporal steps, and average action steps. The results demonstrate that our framework enables efficient and effective collaboration among heterogeneous robots, achieving high success rates across all tasks. Notably, the use of LLMs allows for flexible and decentralized control, as each robot can autonomously make decisions based on its observations and communications, without relying on a centralized controller.

The main contributions of our work are summarised as follows:
\begin{itemize}
    \item The paper introduces \texttt{MHRC}, a novel framework that leverages LLMs for decentralized collaboration among heterogeneous multi-robot systems.
    \item Leveraging a diverse set of feedback, we developed a tailored replanning mechanism for different types of robots. The mobile manipulation robot can dynamically adjust its base position to optimize grasping conditions, while the manipulation robot demonstrates an enhanced understanding of task requirements, ensuring accurate and reliable execution of actions.
    \item We have provided a PyBullet-based simulation benchmark, which encompasses three distinct room layout scenarios and three diverse operational tasks.
    \item We provide comprehensive evaluations demonstrating that our method achieves high success rates and efficient collaboration in complex tasks.
\end{itemize}

\section{Related Work} 

\subsection{LLM for Robotics}

Integrating LLMs into robotic task planning has led to significant advancements in skill learning, control performance, collaboration, human-robot interaction, and navigation~\cite{jansen2020visuallygroundedplanningvisionlanguage,sharma2022skillinductionplanninglatent,huang2022language,li2022pre,ahn2022can,zeng2022socraticmodelscomposingzeroshot}.
LLMs are increasingly used to understand high-level instructions and generate low-level actions that robots can execute directly from textual inputs~\cite{10161317,10160591,ahn2022can}. SayCan~\cite{ahn2022can} combines pre-trained skills with value functions to select appropriate actions for embodied tasks. Code as Policies (CaP)~\cite{10160591} uses structured code generation for robot control, while VoxPoser~\cite{huang2023voxposercomposable3dvalue} employs 3D Value Maps and LLM-generated code for manipulation tasks, with re-planning capabilities in case of failure. Inner Monologue~\cite{huang2022innermonologueembodiedreasoning} improves decision-making by using natural language feedback to guide task planning, and ProgPrompt~\cite{10161317} enables plan generation in diverse environments. Recent methods decompose high-level instructions into executable steps using pre-trained LLMs, often in a zero-shot manner~\cite{huang2022language,ahn2022can}.
Huang et al.~\cite{huang2022language} use GPT-3~\cite{brown2020language} and Codex~\cite{chen2021evaluating} to generate action plans for embodied agents, translating each step into an executable action with the help of the Sentence-RoBERTa model~\cite{liu2019roberta,reimers2019sentence}. The methods discussed above highlight the substantial potential of LLMs in advancing robot task planning. Our work is to use LLMs to build a framework that can realize multi-heterogeneous robot collaboration tasks.

\subsection{Multi-agent Collaboration, Communication, and Motion Planning}

The study of multi-agent systems (MAS) has a rich history~\cite{stone2000multiagent,351231}. MAS can be classified into two types: homogeneous, where agents share similar characteristics, and heterogeneous, where agents have diverse capabilities~\cite{dorri2018multi}.

\subsubsection{Homogeneous Agents}

Sampling-based methods and their various algorithmic improvements have been proposed are a widely used approach in Homogeneous Agents~\cite{karaman2011samplingbasedalgorithmsoptimalmotion,7354289}. Li et al.~\cite{7902130} tackled the optimal synchronization in homogeneous multi-agent systems using an actor-critic neural network and least squares to approximate the control policy and value function. Zhang et al.~\cite{8458223} use a non-policy reinforcement learning algorithm using a single critic neural network to compute each agent's optimal control policy. CoELA~\cite{zhang2023building} leverages LLMs to establish a collaborative framework for homogeneous agents, facilitating cooperative exploration, transport, and communication between the two agents. Guo et al.~\cite{guo2024embodied} places greater emphasis on the organizational framework of LLM agents, investigating how to harness the potential of LLMs to develop more effective collaboration strategies.

\subsubsection{Heterogeneous Agents}

Compared to homogeneous agents, heterogeneous agents present different challenges, including the heterogeneity of agents, limited view of the environment, and the dynamicity of the multi-agent system (MAS) or environment~\cite{1545539}. Several heterogeneous robot combinations have already been proposed, including aerial-ground collaboration, main-picket collaboration, and humanoid-quadruped collaboration~\cite{9341023,9562042,6907527,9340688}. Liu et al.~\cite{liu2024leveraging} applied LLMs to ad hoc and original heterogeneous robots to generate reasonable collaboration strategies. Zhao et al.~\cite{Zhao_2022} developed coordination strategies for robots operating with asymmetric information and varying levels of influence. Haldane et al.~\cite{6907527} address joint locomotion and perception tasks using legged robots of different sizes and capabilities.

\subsection{LLM with closed-loop feedback for robot replanning}

TREE-PLANNER~\cite{hu2024treeplannerefficientcloselooptask} optimizes the closed-loop feedback process through plan sampling, action tree construction, and grounded decision-making. Huang et al.~\cite{huang2022inner} presented a general formulation of Inner Monologue that combines different sources of environmental feedback with methods fusing LLM planning with robotic control policies. HiCRISP~\cite{ming2023hicrisp} allows robots to identify and correct errors at each step of task execution. The re-planning for closed-loop feedback in COME-robot systems~\cite{zhi2024closed} for mobile robots has been primarily focused on the robotic arm, without fully exploiting the inherent mobility of the robot. This oversight limits the robot's effectiveness in tasks such as grasping and transporting objects. Compared to these methods, \texttt{MHRC} emphasizes decision-making and planning tasks for multiple heterogeneous robots by integrating closed-loop feedback specifically associated with robotic capabilities with LLMs.

\section{Method}
\subsection{Preliminary Information}

We formulate this problem as a decentralized partially observable Markov decision process (DEC-POMDP) to facilitate collaborative tasks among heterogeneous multi-robots. Specifically, a DEC-POMDP is defined as $(n, \mathcal{S}, \{\mathcal{A}_i\}, \{\mathcal{O}_i\}, \mathcal{P}, \mathcal{T}, \mathcal{R}, \gamma, \mathcal{H})$, where $n$ is the number of heterogeneous robots in the task; $\mathcal{S}$ denotes a finite set of states; $\mathcal{A}_i = A_i^K \times A_i^C$ represents the action set for robot $i$, including a finite action set $ A_i^K$ determined by the robot's individual capabilities and a communication action $A_i^C$ used to exchange messages with other robots; $\mathcal{O}_i = O_i^K \times O_i^C$ signifies the observation set for robot $i$, encompassing both the observation set $O_i^K$ derived from the robot's perceptions and the message set $O_i^C$ received from other robots; $\mathcal{P}(s'|s, a)$ and $\mathcal{T}(o|s')$ are the state transition probabilities and conditional observation probabilities respectively; $\mathcal{R}$ is the reward function; $\gamma$ is the discount factor in $(0, 1)$; $\mathcal{H}$ is a finite planning horizon. 

It is important to emphasize that our research centers on optimizing the utilization of LLMs for executing heterogeneous multi-robot collaborative tasks, which fall under the domain of robot task and motion planning(TAMP). Consequently, our approach incorporates a description function $f$ that translates the semantics and observations related to the robotic tasks into natural language prompts $l_i^t = f_i(o_i^t), o_i^t \in \mathcal{O}_i$ and does not directly introduce detection-related algorithms within our framework.

\subsection{Multi-Heterogeneous Robot Collaboration with LLMs}

\subsubsection{Observation Module} \label{section:obsm}

This module encompasses all the state information necessary for robot tasks and motion planning. It mainly includes the following three parts:

\textbf{\textit{Scene Graph}}: This part presents structured information about the environment. The scene graph encompasses the positions and orientations of key furniture items. For each piece of furniture, at least one navigation target point (robot's target pose) is defined for ``\textit{navigate}" action. This work is similar to~\cite{habitatrearrangechallenge2022, gu2022multi, makhal2018reuleaux}. In addition to this static information, the scene graph incorporates dynamic updates, including the open/closed status for furniture that requires manipulation and the integration of newly discovered objects, along with updates to their positions and orientations.

\textbf{\textit{Messages Information}}: To achieve effective collaboration in completing tasks, the robots must share information and request assistance as needed. This coordination is enabled through a structured communication process. The experiment begins with the manipulation robot (Bob) initiating two types of requests to the mobile manipulation robot (Alice): one for exploring the environment to locate task-specific objects, and another for transporting objects on the table that are beyond Bob’s operational range. Alice will delegate certain exploration tasks to the mobile robot (David) based on her specific mission requirements. Upon locating the necessary objects, David will relay their positions back to Alice. Notably, all communication occurs via prompts, effectively guiding the LLMs to generate responses autonomously.

\textbf{\textit{Robot Information}}: This section includes the different capabilities of the robot and the corresponding status information. For robots equipped with navigation capabilities, the status includes the robot's pose information; for robots with manipulation capabilities, the status encompasses the gripper's open or closed state, the name of the grasped object, and the maximum grasping range.
    
\subsubsection{Memory Module}

In the context of our work, the tasks are designed as long sequences. As the LLMs are queried multiple times, they may lose track of prior tasks, making it essential to retain past decision contexts. The memory module is designed to capture three key components:

\textbf{\textit{feedback history}}: This section primarily stores limited feedback generated from the robot's past interactions with the environment. We have designed various types of textual feedback for our tasks, covering both success and failure cases. Examples include feedback on the successful execution of atomic actions, updates on the current task status, failure to navigate near the target point, inability to grasp an object due to the suction gripper being too distant, failure to meet task constraints, and failure to satisfy task requirements, among others. For a more comprehensive descriptions of the feedback, please refer to Tab.~\ref{tab:feeb_description}.

\begin{figure}[t]
    \centering
    \includegraphics[width=\columnwidth]{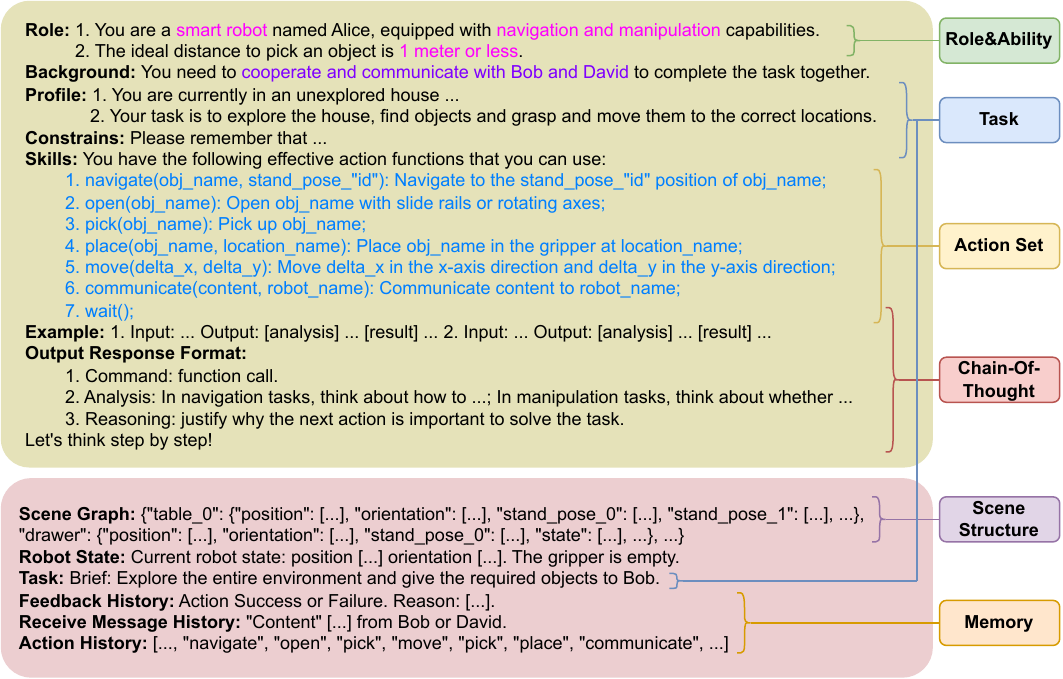}
    \caption{The key prompts for the work are divided into two components: the system prompt and the user prompt. The system prompt(top) remains constant throughout the task, while the user prompt(bottom) dynamically evolves in response to the progression of the task. The figure uses a mobile manipulation robot as an example to illustrate prompt design. For prompt design of other robots and more detailed content, please refer to the appendix~\ref{appendix:prompt}.}
    \label{fig:prompts}
\end{figure}

\textbf{\textit{received message history}}: This part primarily archives messages previously received by the robot from other robots. For detailed information regarding message types, please refer to~\ref{section:obsm}, and for message formats, please refer to Fig.~\ref{fig:prompts}

\textbf{\textit{action history}}: This component primarily archives the sequences of decision-making actions previously executed by the robot, with both types and formats fully adhering to the designed action set~\ref{section:planm}.

In addition, we propose to mark the most recent feedback and received messages with additional tags, as the latest developments often have a more significant impact on the strategy. This approach ensures that LLMs can prioritize recent information while maintaining a comprehensive understanding of the overall context.

\begin{figure}[t]
    \centering
    \includegraphics[width=\columnwidth]{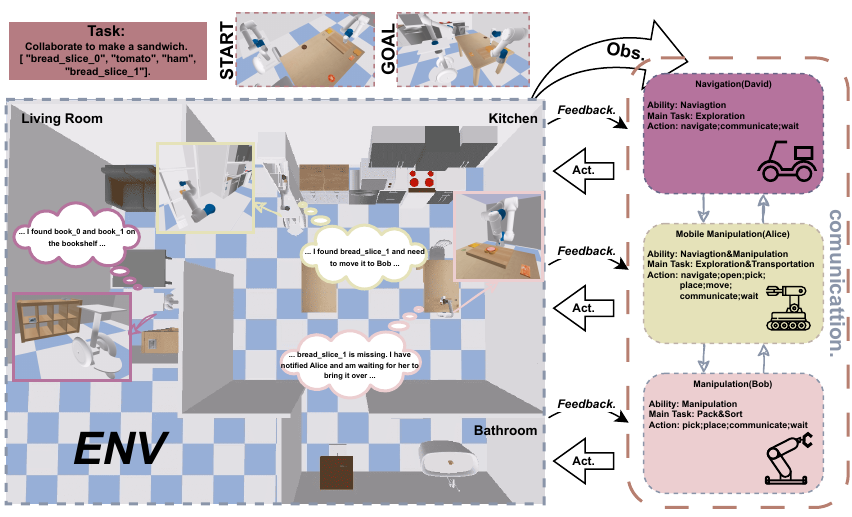}
    \caption{The figure presents the overall workflow using a representative example. The "Start" and "Goal" denote the initial and target states of a task, respectively. Each robot autonomously makes decisions, plans, and executes atomic actions based on its local observations and communications received from other robots. The robots continuously replan their actions in response to environmental feedback. Through coordinated collaboration, multiple heterogeneous robots work together to accomplish complex, long-sequence tasks.}
    \label{fig:overview}
\end{figure}

\subsubsection{Planning Module} \label{section:planm}

We employ LLMs to simulate role-playing for various types of robots, guiding task planning through a CoT framework. This process involves selecting and executing one action from a predefined list at a time. After each action, feedback is obtained, and based on this feedback, along with historical data, the robot autonomously determines the next subtask or adjusts the plan. This multi-turn interaction enables robots to collaborate effectively in completing long-horizon tasks.

In mobile manipulation scenarios, identifying an optimal base position is critical, as it significantly influences the success of grasping actions~\cite{shao2024moma}. To address this, we propose a strategy that combines the $``navigate"$ and $``move"$ actions to accomplish the task. For instance, when grasping an apple from a table, the robot first selects a navigation target from the scene graph and moves near the table. It then evaluates the validity of the robotic arm’s initial and target configurations relative to the apple’s position and attempts to grasp the object. If the attempt fails, the robot selects alternative target points closer to the apple, navigates to those positions, and reattempts the action. Should the task remain incomplete, the robot refines its position using the $``move"$ action, adjusting its base location according to the relative x and y coordinates between the base and the apple.

For tabletop manipulation tasks, it is essential to evaluate the success of pick-and-place actions and ensure that the outcomes align with the task's overall objectives following the execution of the planned motions~\cite{ming2023hicrisp}.

\begin{table*}[h]
\centering
\caption{Feedback Description}
\label{tab:feeb_description}
\renewcommand{\arraystretch}{1.0} 
\resizebox{2.0\columnwidth}{!}{
\begin{tabular}{>{\centering\arraybackslash}m{2.5cm} 
                >{\centering\arraybackslash}m{14.0cm} 
                }
\toprule
\textbf{Type} & \textbf{Description} \\
\midrule
\rowcolor[gray]{.9} \multicolumn{2}{c}{Feedback on successful actions} \\
\midrule
Navigation Success & Provide feedback confirming successful arrival at the designated target point. Additionally, for furniture items such as tables that do not necessitate opening, include information on the types of objects placed on their surfaces. \\
\midrule
Open Success & Provide feedback confirming the successful opening to the target object, along with detailed information regarding the types and positions of items contained within it. \\
\midrule
Move Success & Provide feedback indicating successful displacement by the specified distances along both the x-axis and y-axis. \\
\midrule
Pick Success & Return feedback that the target object has been successfully picked. \\
\midrule
Place Success & Provide feedback on the successful placement of the target object at the specified location. \\
\midrule
\rowcolor[gray]{.9} \multicolumn{2}{c}{Feedback on failed actions} \\
\midrule
Navigation Failed & (1) The starting or ending point of global path planning is deemed invalid if it falls on an obstacle or exceeds the map boundary; (2) The target object for navigation is considered invalid if it does not exist in the scene graph or does not support navigation; (3) A discrepancy greater than an acceptable threshold between the robot's current pose and the target pose can lead to failure.\\
\midrule
Open Failed & (1) The target object is either already in an open state or cannot be opened; (2) The target object is positioned beyond the operational range of the robot. \\
\midrule
Move Failed & The feedback type aligns with "navigation failed" due to the invocation of the API responsible for the \textit{navigate} command. \\
\midrule
Pick Failed & (1) The gripper is grabbing with other objects; (2) The scene graph lacks information about the object to be grasped; (3) The initial or target state is invalid during the verification of the robot arm planning algorithm; (4) The distance between the robot arm’s end effector and the target object exceeds the allowable threshold. Return this distance, and for mobile manipulation robots, also provide the relative distance between the base and the target object along the x and y axes. \\
\midrule
Place Failed & (1) The gripper is empty; (2) The object to be placed does not match the object currently being grasped; (3) The object to be placed has not been placed at the target location. \\
\midrule
\rowcolor[gray]{.9} \multicolumn{2}{c}{Common feedback} \\
\midrule
Target Task Status & Only available for manipulation robots. (1) In the "pack objects" task, the feedback refers to the types of objects present in the tray; (2) In the "sort solids" task, it provides information regarding the shapes and colors of solids placed on panels of different colors; (3) In the "make sandwich" task, the feedback describes the types of ingredients arranged on the cutting board from bottom to top. \\
\bottomrule
\end{tabular}}
\end{table*}

\begin{algorithm}[t]
\caption{Multi-Heterogeneous Robot Collaboration}
\begin{algorithmic}[1]
\Require heterogeneous robots $r_0, \cdots, r_i, \cdots, r_n$, planning horizon $\mathcal{H}$
\Require memory buffer $b_i \subset \mathcal{B}$, prompt policy $\pi_i \in \Pi$
\State $\mathcal{O}$ $\leftarrow$ env.reset()
\For{step $t = 0 \  \textbf{to} \  \mathcal{H}$}
    \For{robot $i = 1 \  \textbf{to} \  n$}
        \State $a_i^t = \pi_i(b_i, f_i(o_i^t))$
        \Comment{atomic action $a_i^t \in \mathcal{A}_i$}
        \State $f_i^t, m_i^t$ = env.step($a_i^t$)
        \Comment{limited feedback $f_i^t$ from env after $a_i^t$ is executed and message $m_i^t$ received from other robots $r_{n \neq i}$}
        \State $m_i \leftarrow m_i \cup \{a_i^t, f_i^t, m_i^t\}$ 
    \EndFor
    \State $\mathcal{B} \leftarrow \mathcal{B} \cup b_i$
\EndFor
\end{algorithmic}
\end{algorithm}

\section{Experiments}

\subsection{Environments, Tasks, and Metric}

In this study, we constructed several experimental environments using the BestMan~\cite{ding2023task, ding2023integrating} simulation platform and developed additional functionalities to support the specific tasks of our research. Each environment features a variety of room types and layouts, diverse furniture types and arrangements, and an array of different objects. The first row of Fig.~\ref{fig:first} illustrates the environments corresponding to these three different scenarios, respectively. For the robot navigation algorithm, we use the A*~\cite{Kurzer1057261} algorithm for global path planning without considering local obstacle avoidance~\cite{fox1997dynamic, yu2024pathrl, yu2024ldp}, as it is not the focus of our task. For the robotic arm planning algorithm, we use the sample-based BIT*~\cite{gammell2015batch} algorithm, considering path planning in a static state.

We also designed the following three different tasks:

\begin{itemize}
    \item \textit{Pack Objects}: The objective of this task is to evaluate the robot's fundamental picking and placing capabilities. The robot is provided with a list of objects and must accurately place each object into a designated tray. The objects include \textit{apple}, \textit{fork}, \textit{soap}, \textit{toy duck}, \textit{phone}, \textit{bottle}, \textit{book}, etc.
    \item \textit{Sort solids}: In addition to evaluating the robot's basic pick-and-place capabilities, this task also requires the robot to perform color matching. The robot must accurately place solids, each of a different color, onto the corresponding colored panels on the table. The colors include \textit{red}, \textit{blue}, \textit{pink}, \textit{green}, \textit{yellow}, and \textit{purple}.
    \item \textit{Make Sandwish}: This task further tests the robot's ability to stack objects in a specific order. The robot is tasked with assembling sandwiches of varying flavors based on a given menu, requiring it to sequentially stack the sandwich ingredients. The ingredients include \textit{bread slices}, \textit{ham}, \textit{bacon}, \textit{tomato}, \textit{cucumber}, \textit{cheese}, and \textit{beef patties}.
\end{itemize}

Additionally, in the three types of tasks outlined above, the mobile robot (David) is responsible for the efficient exploration of the environment. The mobile manipulation robot (Alice), beyond exploration, is tasked with opening furniture and transporting objects. The manipulation robot (Bob) handles operations on the tabletop. Alice is not capable of directly executing the final manipulation tasks; all tabletop operations are performed by Bob.

We introduce four metrics to evaluate the performance of different methods:

\begin{itemize}
    \item \textit{Success rate (Succ)}: The rate of episodes in which the robot completes the full task, meaning all objects are placed in the correct positions.
    \item \textit{Partial success rate (PS)}: The average ratio of correctly placed objects to the total number of objects in each episode. The rationale for designing this metric is that, within a task episode, an increase in the number of objects and the length of the sequence generally makes the task more challenging. However, with more objects, the impact of a single failure tends to be smaller. Conversely, in tasks involving fewer objects, the consequences of a single failure are more significant.
    \item \textit{Average temporal steps (TS)}: the average number of temporal steps in all episodes.
    \item \textit{Average action steps (AS)}: the average number of action steps in all episodes, excluding \textit{wait} actions. The robot's capability to execute \textit{wait} actions at optimal moments can significantly reduce energy consumption, demonstrating that the policy has a comprehensive understanding of the entire task.
\end{itemize}

\subsection{Experiments on simulation scenarios}

\begin{table}[htbp]
\centering
\caption{Evaluation Results on Our Environment and Tasks}
\label{tab:main}
\renewcommand{\arraystretch}{1.2} 
\resizebox{\columnwidth}{!}{
\begin{tabular}{>{\centering\arraybackslash}p{3.2cm} 
                >{\centering\arraybackslash}p{1.0cm} 
                >{\centering\arraybackslash}p{1.0cm} 
                >{\centering\arraybackslash}p{1.0cm} 
                >{\centering\arraybackslash}p{1.0cm}
                }
\toprule
\textbf{Methods} & \textbf{Succ(\%)}$\uparrow$ & \textbf{PS(\%)}$\uparrow$ & \textbf{TS}$\downarrow$ & \textbf{AS}$\downarrow$ \\
\midrule
\rowcolor[gray]{.9} \multicolumn{5}{c}{Make Sandwich} \\
MHRC $w.$ GPT-3.5-turbo & 0.00 & 0.14 & 50.00 & 41.22 \\
MHRC $w.$ Llama-3.1-8B & 0.00 & 0.12 & 50.00 & 46.00 \\
MHRC $w.$ GPT-4o & \textbf{0.75} & \textbf{0.81} & \textbf{38.89} & \textbf{31.11} \\
\rowcolor[gray]{.9} \multicolumn{5}{c}{Sort Solids} \\
MHRC $w.$ GPT-3.5-turbo & 0.00 & 0.17 & 50.00 & 39.89 \\
MHRC $w.$ Llama-3.1-8B & 0.00 & 0.14 & 50.00 & 44.33 \\
MHRC $w.$ GPT-4o & \textbf{0.83} & \textbf{0.88} & \textbf{34.56} & \textbf{27.33} \\
\rowcolor[gray]{.9} \multicolumn{5}{c}{Pack Objects} \\
MHRC $w.$ GPT-3.5-turbo & 0.00 & 0.25 & 50.00 & 37.67 \\
MHRC $w.$ Llama-3.1-8B & 0.00 & 0.20 & 50.00 & 40.33 \\
MHRC $w.$ GPT-4o & \textbf{0.92} & \textbf{0.96} & \textbf{32.67} & \textbf{25.00} \\ 
\bottomrule
\end{tabular}}
\end{table}

\subsubsection{Comparative experiments with different LLMs}

In our framework, we evaluated various LLMs, including the closed-source models GPT-3.5-turbo and GPT-4o, as well as the open-source Llama-3.1-8B. Tab.~\ref{tab:main} summarizes their performance across multiple metrics. Each task was tested in three distinct scenarios, with four trials per scenario. The number of objects per trial was set to 3, 4, 5, and 6, respectively, creating task sequences of varying lengths. We ensured that each task involved objects requiring exploration and transportation by the robots, as well as objects positioned on a table but beyond the reach of the tabletop robots. The maximum step count for each experiment was limited to 50, and all models were tested with a temperature setting of 0.5. Experimental results showed that, aside from GPT-4o, the other models struggled to complete full tasks. They were only able to successfully grasp objects near the robotic arm and frequently made logically inconsistent decisions, such as issuing an open command far from the refrigerator.

\subsubsection{Ablation study}

Our multi-heterogeneous robot collaboration framework comprises several key modules, with Tab.~\ref{tab:ablation} detailing the performance outcomes when various components are ablated. The experimental results highlight the necessity of both feedback and history, as they play a pivotal role in determining task execution success. Without access to history information, the model may lose track of prior actions, leading to behaviors such as re-exploring the same location or forgetting previously received requests. This results in unnecessary steps due to redundant communication. Additionally, the manipulation robot often focuses only on nearby objects, failing to collaborate effectively on objects that require coordination.

\begin{table}[htbp]
\centering
\caption{Evaluation Results on Our Environment and Tasks}
\label{tab:ablation}
\renewcommand{\arraystretch}{1.2} 
\resizebox{\columnwidth}{!}{
\begin{tabular}{>{\centering\arraybackslash}p{3.0cm} 
                >{\centering\arraybackslash}p{1.0cm} 
                >{\centering\arraybackslash}p{1.0cm} 
                >{\centering\arraybackslash}p{1.0cm} 
                >{\centering\arraybackslash}p{1.0cm}
                }
\toprule
\textbf{Methods} & \textbf{Succ(\%)}$\uparrow$ & \textbf{PS(\%)}$\uparrow$ & \textbf{TS}$\downarrow$ & \textbf{AS}$\downarrow$ \\
\midrule
\rowcolor[gray]{.9} \multicolumn{5}{c}{Make Sandwich} \\
MHRC $w.o.$ feedback & 0.00 & 0.08 & 50.00 & 48.56 \\
MHRC $w.o.$ history & 0.00 & 0.48 & 50.00 & 46.89 \\
MHRC $w.o.$ mobile robot & 0.69 & 0.78 & 44.67 & 36.33 \\
MHRC $w.$ GPT-4o & \textbf{0.75} & \textbf{0.81} & \textbf{38.89} & \textbf{31.11 }\\
\rowcolor[gray]{.9} \multicolumn{5}{c}{Sort Solids} \\
MHRC $w.$ feedback & 0.00 & 0.10 & 50.00 & 46.22 \\
MHRC $w.$ history & 0.00 & 0.56 & 50.00 & 47.44 \\
MHRC $w.o.$ mobile robot & 0.81 & 0.84 & 40.50 & 31.17 \\
MHRC $w.$ GPT-4o & \textbf{0.83} & \textbf{0.88} & \textbf{34.56} & \textbf{27.33} \\
\rowcolor[gray]{.9} \multicolumn{5}{c}{Pack Objects} \\
MHRC $w.$ feedback & 0.00 & 0.14 & 50.00 & 46.00 \\
MHRC $w.$ history & 0.00 & 0.61 & 50.00 & 48.33 \\
MHRC $w.o.$ mobile robot & \textbf{0.92} & 0.94 & 37.00 & 28.33 \\
MHRC $w.$ GPT-4o & \textbf{0.92} & \textbf{0.96} & \textbf{32.67} & \textbf{25.00} \\ 
\bottomrule
\end{tabular}}
\end{table}

The absence of feedback proves even more detrimental, causing a significant decrease in task success rates and the emergence of various errors. For example, the mobile manipulation robot might not detect a failed grasp attempt and proceed with subsequent actions regardless, even sending incorrect status updates to the manipulation robot, claiming successful placement of objects on the table. Similarly, the manipulation robot may limit its actions to nearby objects, ignoring the broader task requirements. Furthermore, the robots operating within the framework without feedback and history exhibited significantly higher AS values, suggesting that these systems lacked a comprehensive understanding of the overall task, resulting in excessive redundant actions.

While theoretically, Alice and Bob alone are sufficient to complete all the experimental tasks, we explored the impact of removing David from the framework to assess the resultant performance. As demonstrated by the experimental results in Tab.~\ref{tab:ablation}, David plays a crucial role in enhancing task efficiency. 

\section{Conclusions and Future Work}

In this paper, we present \texttt{MHRC}, a novel framework that leverages LLMs for decentralized collaboration among heterogeneous multi-robot systems. We designed tailored feedback mechanisms for different robot types, enabling them to replan based on real-time feedback, thus improving task success rates. Our approach demonstrated effective decision-making and planning, as validated through a series of simulated tasks.

In future work, two aspects can be further explored: (1) The integration of additional heterogeneous robots to address more complex environments and tasks, such as deploying legged robots to navigate uneven terrains like stairs, and drones to explore aerial targets. (2) Although the current framework can complete tasks within a limited time, it remains inefficient, with redundant actions. Thus, developing a more optimized collaboration framework, along with enhanced CoT approaches, will facilitate more efficient robot cooperation.

\section*{APPENDIX}

\subsection{Detailed prompt templates} \label{appendix:prompt}

Due to the extensive content of the prompt template and the constraints of page space, please visit our website for detailed information.~\url{https://github.com/WenhaoYu1998/ICRA25/blob/main/ICRA25\_MHRC\_appendix.pdf} 

\bibliographystyle{IEEEtran}
\bibliography{IEEEexample}

\end{document}